\pdfoutput=1
\documentclass[11pt]{article}

\usepackage[]{EMNLP2023}

\usepackage{times}
\usepackage{latexsym}

\usepackage[T1]{fontenc}
\usepackage[utf8]{inputenc}

\usepackage{microtype}
\usepackage{bbm}

\usepackage{inconsolata}

\usepackage[utf8]{inputenc} % allow utf-8 input
\usepackage[T1]{fontenc}    % use 8-bit T1 fonts
\usepackage{hyperref}       % hyperlinks
\usepackage{url}            % simple URL typesetting
\usepackage{booktabs}       % professional-quality tables
\usepackage{amsfonts}       % blackboard math symbols
\usepackage{nicefrac}       % compact symbols for 1/2, etc.
\usepackage{microtype}      % microtypography
\usepackage{xcolor}         % colors
\usepackage{soul}
\usepackage{amsmath}
\usepackage{graphicx}
\usepackage{tikz}
\usepackage{pgfplots}
\usepackage[normalem]{ulem}
\usepackage{cleveref}
\usepackage{hyperref}       %
\usepackage{url}            %
\usepackage{booktabs}       %
\usepackage{amsfonts}       %
\usepackage{nicefrac}       %
\usepackage{microtype}      %
\usepackage{xcolor}         %
\usepackage{xspace}
\usepackage{amsmath}
\usepackage{amssymb}
\usepackage{MnSymbol}
\usepackage{pifont}
\usepackage{mathtools}
\usepackage{amsthm}
\usepackage{bm}
\usepackage{wrapfig}
\usepackage{multirow}
\usepackage[normalem]{ulem}
\usepackage{tikz}
\usepackage{tikzsymbols}
\usepackage{adjustbox}
\usetikzlibrary{arrows.meta}
\usetikzlibrary{shapes.misc, positioning}
\usepackage{pgfplots}
\usepgfplotslibrary{groupplots}
\pgfplotsset{compat = 1.5}
\usepackage{cleveref}
\usepackage{tabularx}
\usepackage{graphicx}
\usepackage{caption, subcaption}
\usepackage{paralist}
\usepackage{tabularx}
\usepackage{ulem}
\usepackage{xfakebold}
\usepackage{listings}
\usepackage{color}
\usepackage{selectp}
\usepackage{pict2e}
\usepackage{adjustbox}
\usepackage{colortbl}

\newcommand{\model}{{\texttt{TRAMS}}\xspace}

\newcommand{\enwik}{\texttt{enwik8}\xspace}
\newcommand{\wt}{\texttt{WikiText-103}\xspace}

\def\v1{{\vec{\bm{\mathbbm{1}}}}}

\title{ \model: Training-free Memory Selection for Long-range Language Modeling}

\author{
Haofei Yu\textsuperscript{$\heartsuit$\thanks{\ \ Work done during internship at Tencent AI Lab.}}, 
Cunxiang Wang\textsuperscript{$\clubsuit$\thanks{\ \ Co-first Author.}}, 
Yue Zhang\textsuperscript{$\clubsuit$},
Wei Bi\textsuperscript{$\diamondsuit$\thanks{\ \ The correponding author.}}
\\
\textsuperscript{$\heartsuit$}Language Technologies Institute, Carnegie Mellon University, USA\\
\textsuperscript{$\clubsuit$}School of Engineering, Westlake University, China 
\textsuperscript{$\diamondsuit$} Tencent AI Lab, China\\
  {\texttt{haofeiy@cs.cmu.edu}, \texttt{\{wangcunxiang, zhangyue\}@westlake.edu.cn}}, \\ 
   {\texttt{victoriabi@tencent.com}}
}

\begin{document}
\maketitle
\begin{abstract}
The Transformer architecture is crucial for numerous AI models, but it still faces challenges in long-range language modeling. Though several specific transformer architectures have been designed to tackle issues of long-range dependencies, existing methods like Transformer-XL are plagued by a high percentage of ineffective memories. In this study, we present a plug-and-play strategy, known as \textbf{TRA}ining-free \textbf{M}emory \textbf{S}election (\model), that selects tokens participating in attention calculation based on one simple metric. This strategy allows us to keep tokens that are likely to have a high attention score with the current queries and ignore the other ones. We have tested our approach on the word-level benchmark (\wt) and the character-level benchmark (\enwik), and the results indicate an improvement without having additional training or adding additional parameters.
\end{abstract}

\section{Introduction}
Transformer-based models~\cite{bert, roberta, T5, albert, gpt} have achieved remarkable performance over the past few years. The key component of these model architectures is the attention mechanism~\cite{vanilla-transformer}. However, the original attention design struggles to efficiently handle long sequences, which becomes particularly problematic in scenarios such as document-level translation~\citep{werlen2018document, kim2019and} and large-scale text generation~\citep{zhou2023recurrentgpt}, as its time and space computation costs increase quadratically with the sequence length \citep{tay2022efficient}. The primary factor for this elevated computational complexity can be traced back to the multiplication between queries and keys used in the attention module. In general, the time complexity for calculation is $\mathcal{O}(N^2d)$ if a transformer model with $d$ dimensions is set up with an input consisting of $N$ tokens.

\begin{figure}[t]
  \centering
  \includegraphics[width=0.47\textwidth]{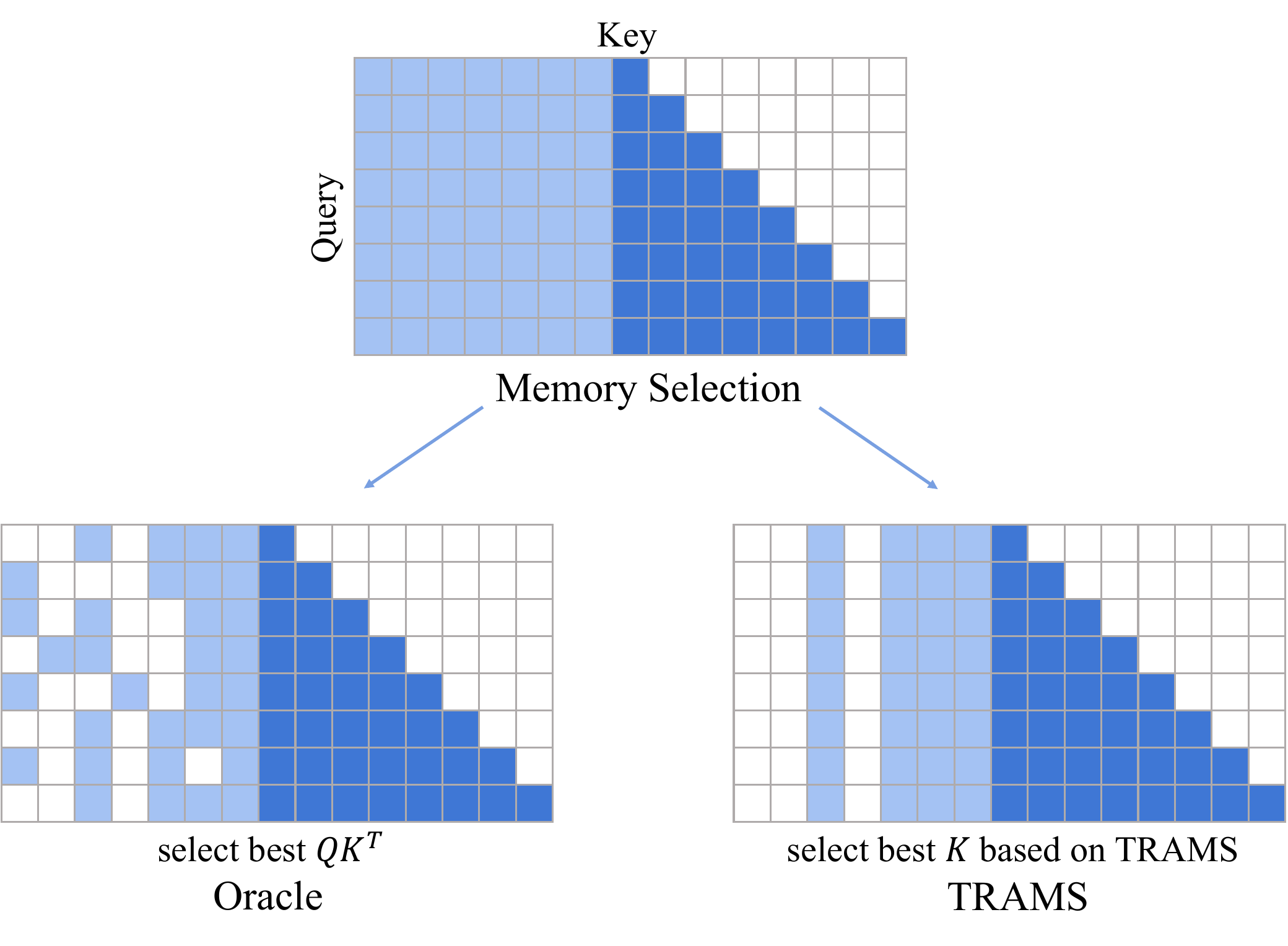}
  \caption{Two memory selection methods: For oracle, it selects memories with the highest attention scores after computing $QK^\top$. For \model, it selects important key/value pairs that are independent of queries based on our self-defined metric before computing $QK^\top$.}
  \label{fig:my_image}
  \vspace{-3mm}
  \label{fig:transformer-xl} 

\end{figure}

To tackle this computation bottleneck, numerous efforts have been made. The first line of work is to find a new efficient expression to compute the attention score. Despite the advancements made, these methods often compromise performance, thus paving the way for alternative solutions. Efficient architectures that provide an approximate expression of attention have been explored widely~\citep{wang2020linformer, rfa, peng2022abc,choromanski2021hybrid, zheng2022efficient, zheng2022linear}. The second line of work is to keep the calculation expression the same and use an external structure like hash function~\citep{kitaev2019reformer, daras2020smyrf}, clustering~\citep{routing-transformer, vyas2020fast} and memory selector~\citep{pietruszka2022sparsifying, transformer-xl, bertsch2023unlimiformer, expire-span, adaptive-span, child2019generating} to find the suitable subset of queries and keys in the long sequence for attention calculation.

Our work falls into the second category, in which we propose a training-free memory selection mechanism to select suitable tokens for attention computation. Specifically, we focus on pushing Transformer-XL~\citep{transformer-xl} architecture to a better position by selecting higher-quality tokens inside its memory. Based on our initial investigation, we construct a  memory subset by selecting 50\% of the memories with the largest attention values and maintaining the same performance. It indicates that \textbf{a large portion of information in memory is not fully utilized}. This motivates us to explore better methods to optimize memory usage.

Illustrated in Figure~\ref{fig:transformer-xl}, we propose a \textbf{TRA}ining-free \textbf{M}emory \textbf{S}election method (\model) that can be directly plugged into memory-based long-range language models and reduces the time complexity of computing attention matrix. Through experiments on two language modeling benchmark datasets, namely word-level \wt~\citep{wikitext103} and character-level \enwik~\citep{enwik8}, we achieve an improvement in the model's performance, as demonstrated by a 0.19 perplexity (ppl) drop in \wt and a 0.017 reduction in bits-per-character (bpc) in \enwik.

To our knowledge, we are the first to design a training-free memory selection method based on Transformer-XL architecture.\footnote{Source code for this paper is available at \href{https://github.com/lwaekfjlk/TRAMS}{https://github.com/lwaekfjlk/TRAMS}.}

\section{Method}
\subsection{Problem Definition}
We use $\bm{h} \in \mathbb{R}^{N \times d}$ to represent the input hidden states for the attention module, $\bm{o} \in \mathbb{R}^{N \times d}$ to represent the output hidden states for the attention module, $\bm{m} \in \mathbb{R}^{M \times d}$ to represent the memory hidden states used in the attention calculation. We use $W_Q$, $W_K$, $W_V$ to represent the trainable projection matrix in the attention module. We define $d$ for the dimension of the model, $M$ for the memory size, and $N$ for the input size. The attention calculation process can be formally written as
%\begin{equation}
$\bm{o} = \text{Attn}(\bm{h},\bm{m})$.
%\end{equation}

With the above annotations, the problem of memory selection can be defined as choosing a subset of hidden states memory $\bm{\tilde{m}}$ from the memory $\bm{m}$ that brings the minimum difference to the transformer layer output but with a smaller memory size.
\begin{align}
 {\tilde{\bm{m}}}^* & = \mathop{\arg\min}_{\tilde{\bm{m}} \subset \bm{m}}\|\text{Attn}(\bm{h},\tilde{\bm{m}})-\text{Attn}(\bm{h}, {\bm{m}})\|
\end{align}

\subsection{Attention Reformulation}
\paragraph{Standard Attention} In a memory-augmented language model, the standard attention mechanism~\citep{vanilla-transformer} between input hidden states and memory hidden states can be written as:
\begin{equation}
\text{Attn}(\bm{h}, \bm{m}) = \text{softmax}(\frac{QK^\top}{\sqrt{d}})V
\end{equation}
where $Q=\bm{h}W_Q$ is the product of target token hidden states $\bm{h}$ and query projection matrix $W_Q$; $K=\bm{m}W_K$ is the product of memory token hidden states $\bm{m}$ and key projection matrix $W_K$; $V=\bm{m}W_V$ is also the product of memory token hidden states $\bm{m}$ and value projection matrix $W_V$.

\paragraph{Unlimiformer Attention} Different from the well-known attention score calculation, Unlimiformer~\citep{bertsch2023unlimiformer} proposed a rewritten way to compute the dot-product part of cross-attention in the encoder-decoder architecture:
\begin{align}
QK^\top& = ({\bm{h}_d W_Q})({\bm{h}_eW_K})^\top \nonumber \\
       & = (\bm{h}_d W_Q W_K^\top) \bm{h}_e^\top
\end{align}
where $\bm{h}_e$ is the encoder hidden state and $\bm{h}_d$ is the decoder hidden state. It allows Unlimiformer to avoid indexing the keys for each head and layer separately and avoid storing values in a separate index from the keys during $k$NN-based searching and retrieval stage, making it more efficient.

\paragraph{\model Attention}
Even though we have no need to store or index any key or value for our method, Unlimiformer attention motivates us to transfer more useful information to keys by reformulating attention and allows us to do more effective memory selection solely based on reformulated keys. We can compute this attention formula in a different order but maintain the same result:
\begin{align}
QK^\top& = ({\bm{h}W_Q})({\bm{m}W_K})^\top \nonumber \\
       & = ({\bm{h}}) ( \bm{m} W_K W_Q^\top)^\top
\end{align}
Thus, we define $Q' = \bm{h}$ as the reformulated query for this attention expression and $K' = \bm{m} W_K W_Q^\top$ as the reformulated keys for attention. With this reformulation, we transfer all attention-related parametric information onto reformulated key vectors.

\subsection{Transformer Hidden Space}
Since $\bm{h}$ is the input of the current transformer layer and also the output of the previous transformer layer, it is the result of the last layer's \texttt{Layernorm} operation. We can define the coordinate-wise average of $\bm{h}$ as $\mu$ and the coordinate-wise standard deviation of $\bm{h}$ as $\sigma$. Expressions can be written as:
\begin{align}
    \mu = \frac{1}{d}\sum_{i=1}^d h_i \approx 0, \;\;
    \sigma = \sqrt{\frac{1}{d}\sum_{i=1}^d (h_i - \mu)^2} \approx 1
\end{align}
Since the mean value for the hidden states $\bm{h}$ is around zero, we can confirm the hidden states vectors are approximately orthogonal to the $\v1$ vector and the L2 norm of hidden states is around $\sqrt{d}$.

\begin{figure}
  \centering
  \includegraphics[scale=0.45]{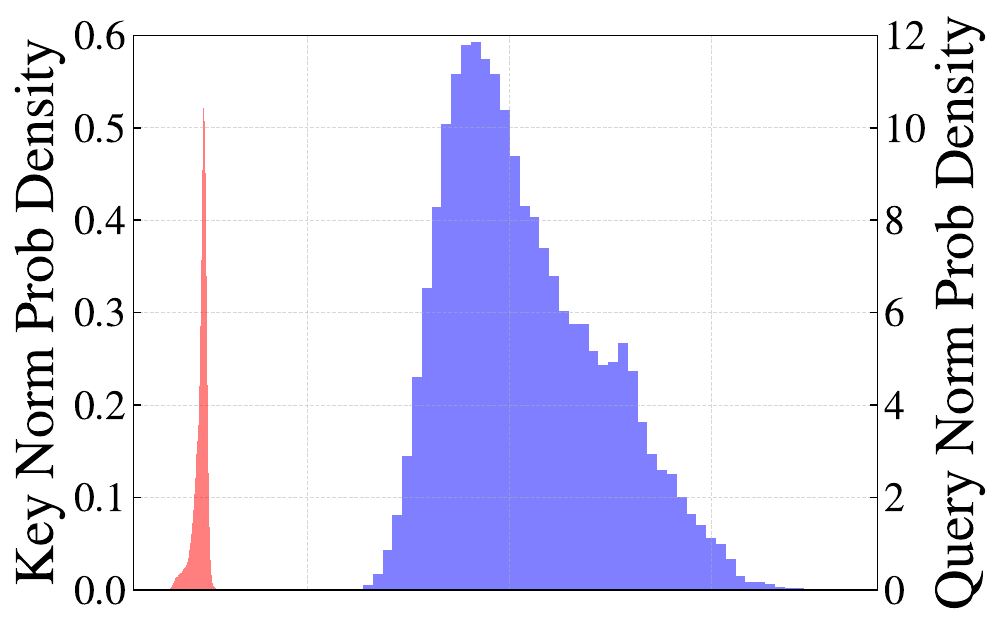}
  \caption{Norm distribution of reformulated $Q'$ and $K'$. The \colorbox{red!30}{red} distribution represents the query norm. The \colorbox{blue!30}{blue} distribution represents the key norm.}
  \label{fig:norm_distribution}
  \vspace{-2mm}
\end{figure}

With this approximation, we can expand our reformulated attention score as:
\begin{align}
Q' K'^\top &= ({\bm{h}}) ( \bm{m} W_K W_Q^\top)^\top \nonumber \\
           &= ||Q'|| \cdot ||K'|| \cdot \cos\left< Q', K' \right> \nonumber \\ 
           &\approx \sqrt{d} \cdot ||K'|| \cdot \cos\left< Q', K' \right>
\end{align}
where $\|Q'\|$ stands the L2 norm for $Q'$ and $\|K'\|$ stands for the L2 norm for $K'$.
Based on Fig \ref{fig:norm_distribution}, we see that reformulated query norm $||Q'||$ has a much sharper distribution compared with key norm $||K'||$, indicating reformulated query norm can be approximated by a constant factor.

\subsection{Training-free Memory Selection (\model)}
Our target for memory selection is to recover the complete attention score with as few memory tokens as possible. This problem is equivalent to finding the subset of memory tokens that have the highest attention scores with queries. We propose a heuristic method to perform token-level selection for each layer and each head based on a memory-independent metric in this section.

There are two crucial components for calculating the attention score after approximating $||Q'||$ with a constant factor: the norm of the reformulated keys $||K'||$ and the angles between the reformulated keys and queries $\arccos\left< Q', K' \right>$, which is proved in \citet{knn-lm}. Commonly, we believe that $\arccos\left< Q', K' \right>$ is the more important factor in general. Yet, if we use the ranking of attention score value for all query and key pairs as ground-truth ranking, based on Fig~\ref{fig:correlation}, we empirically discovered that rankings based on key norms and rankings based on angles produce close Spearman correlation scores when only taking the highest 1\% attention scores into account. Therefore, it indicates that we can rank our memory tokens based on $||K'||$ solely to gain a relatively good performance when we desire top 1\% attention scores with queries in our memories instead of all. 

Additionally, we discovered that relying solely on a large norm isn't sufficient as a constraint. Specifically, keys that are nearer to $\v1$ tend to yield a higher attention score. To address this, we introduce a combined metric: $s = \cos\left< K', \v1 \right>||K'||$. This metric allows us to identify tokens that can produce \textit{high} attention scores when paired with the appropriate query (owing to a high value of $||K'||$) and \textit{low} scores when paired with an unsuitable query (owing to the high level of orthogonality with the query space based on $\cos\left< K', \v1 \right>$). This is due to the near orthogonality to the query space, as indicated by a small angle with $\v1$, which is orthogonal to the query space.

\pgfplotstableread[row sep=\\,col sep=&]{
    n & CodeT5 & RACOON-CBS-F1 & RACOON-CBS-P & RACOON-ROUGE & RACOON-BLEU & RACOON-CODEBLEU & RACOON-CBS-R  \\
    1    &  0.4272  & 0.4573 &  0.4328  & 0.4437 & 0.4370 & 0.4348 & 0.4528 \\
    2    &  0.2845  & 0.3260 &  0.3129  & 0.3190 & 0.2962 & 0.3009 & 0.3138 \\
    3    &  0.2032  & 0.2392 &  0.2316  & 0.2336 & 0.2159 & 0.2147 & 0.2287 \\
    4    &  0.1492  & 0.1779 &  0.1743  & 0.1737 & 0.1549 & 0.1536 & 0.1710\\
    }\ngramdataconala

\pgfplotstableread[row sep=\\,col sep=&]{
    n & CodeT5 & RACOON-CBS-F1 & RACOON-CBS-P & RACOON-ROUGE \\
    1    &  7.32  & 6.34 & 6.66 & 7.00 \\
    3    &  12.62  & 11.40 & 11.98 & 12.36 \\
    5    &  15.68  & 14.49 & 15.17 & 15.51 \\
    10    &  19.98  & 19.18 & 20.02 & 20.10  \\
    30    &  26.40  & 26.36 & 27.91 & 27.15 \\
    50    &  29.11  & 29.38 & 31.06 & 30.18  \\
    100    &  32.61  & 33.48 & 34.35 & 34.78  \\
    }\passkdataconala

\definecolor{color1}{rgb}{0.9804, 0.9176, 0.8275} % 250/255, 234/255, 211/255
\definecolor{color2}{rgb}{0.4824, 0.5451, 0.4353} % 123/255, 139/255, 111/255
\definecolor{color3}{rgb}{0.7882, 0.7529, 0.8275} % 201/255, 192/255, 211/255

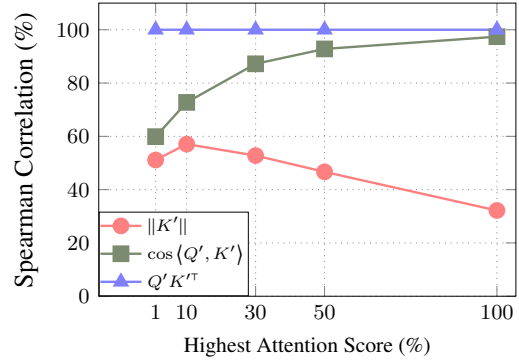
\begin{figure}[t!]
    \vspace{2mm}
    \begin{tikzpicture}[scale=0.90]
      \begin{axis}[
        xlabel={Highest Attention Score (\%)},
        ylabel={Spearman Correlation (\%)},
        xlabel style={font=\footnotesize},
        ylabel style={font=\footnotesize},
        yticklabel style={font=\footnotesize}, 
        xticklabel style={font=\footnotesize}, 
        ylabel near ticks,
        legend style={at={(0,0)},anchor=south west,mark size=1pt,font=\scriptsize,
        inner xsep=1pt, inner sep=0pt},
        legend cell align={left},
        xmin=-10, xmax=100,
        ymin=0, ymax=110,
        xtick={1,10,30,50,100},
        ytick={0,20,40,60,80,100},
          grid = major,
          major grid style={dotted,gray},
          height=5.9cm,
          width=0.48\textwidth,
          enlarge x limits=0.05,
          ]
          
          \addplot[color=red!50!white, solid, mark options={solid, fill=red!50!white, draw=none}, mark=*, line width=1pt, mark size=3pt, ] 
          table [meta index=2]  {
            x   y       label   alignment
            1	51.13	51.13		-50
            10	57.08	57.08		-30
            30  52.78   52.78       -50
            50  46.67   46.67       -50
            100 32.18   32.18		-50
            }; %
            \addlegendentry{$||K'||$}

            \addplot[color=color2, solid, mark options={solid, fill=color2, draw=none}, mark=square*, line width=1pt, mark size=3pt] 
            table [meta index=2]  {
            x   y       label   alignment
            1   59.89   59.89       170
            10	72.74	72.74		170
            30	87.22	87.22		170
            50  92.80   92.80       170
            100 97.40	97.40		50
            };
            \addlegendentry{$\cos\left< Q', K' \right>$}

            \addplot[color=blue!50!white, solid, mark options={solid, fill=blue!50!white, draw=none}, mark=triangle*, line width=1pt, mark size=3pt] 
            table [meta index=2]  {
            x   y       label   alignment
            1	100.0	100.0		30
            10	100.0	100.0		30
            30	100.0	100.0		-30
            50  100.0	100.0		-30
            100 100.0	100.0		-30
            };
            \addlegendentry{$Q'K'^\top$}
      \end{axis}
  
  \end{tikzpicture}
  \vspace{-2mm}
  \caption{Spearman Correlation Score on different ranking metrics with the groundtruth one.}
  \vspace{-6mm}
  \label{fig:correlation} 

\end{figure}

\section{Experiments}
We introduce the compared methods and report the main results and analysis on different attention variants for inference in this section. Datasets details for \wt and \enwik benchmarks and their evaluation metric details are included in Appendix \ref{sec:appendix3}. The details of the model that we built memory selection on can be seen in Appendix \ref{sec:appendix}.

\subsection{Compared Methods}
% We introduce the details of our baseline models for comparison.

\textbf{Transformer+RPE}~\citep{vanilla-transformer}: the vanilla transformer baseline with relative position embedding that is the same as Transformer-XL. Therefore, the only difference between this model and Transformer-XL is the additional memories. More information related to relative position embedding can be seen in Appendix \ref{sec:appendix2}.

\noindent \textbf{Transformer-XL}~\citep{transformer-xl}: a specific-designed architecture for long-range language modeling. It includes relative position embedding and recurrent memories per layer. Memory slots are filled with hidden states from previous time steps.

\subsection{Experimental Settings}
We compare our methods with the Transformer-XL~\citep{transformer-xl} under the same size of memory ($m=200$) for attention calculation. For the input token length $n$ for both models, we keep the same as in \cite{transformer-xl} ($n=64$). Additionally, the memory selection process is performed on a memory pool with the size of $M$. Our model and the Transformer-XL share the model parameters but have different inference strategies.

\subsection{Main Results}
The main results of \wt and \enwik datasets are shown in Table \ref{tab:wt103}. Without additional training or additional parameters, we gain 0.19 improvement in perplexity and 0.017 improvement for bit-per-character with our \model mechanism. We implement $p$-test by inferencing on multiple model checkpoints and prove that our results are significant ($p$ < 0.05).

\begin{table}
\centering
\resizebox{0.42\textwidth}{!}{
\begin{tabular}{lcccc}

\toprule[1.0pt]
\multicolumn{5}{c}{\wt}\\
\midrule[1.0pt]
Model  & M & m & n & PPL ($\downarrow$)\\
\midrule[1.0pt]
Transformer+RPE & - & - & - & 29.14 \\
Transformer-XL & - & 200 & 64 & 24.17 \\
\model & 400 & 200 & 64 & \textbf{23.98} \\
\midrule[1.0pt]
\multicolumn{5}{c}{\enwik}\\
\midrule[1.0pt]
Model  & M & m & n & bpc ($\downarrow$)\\
\midrule[1.0pt]
Transformer+RPE & - & - & - & 1.240 \\
Transformer-XL  & - & 200 & 64 & 1.215 \\
\model & 400 & 200 & 64 & \textbf{1.198}  \\
\bottomrule[1.0pt]
\end{tabular}}
\caption{Model performance on the word-level \wt and the character-level \enwik datasets. }
\vspace{-6mm}
\label{tab:wt103}
\end{table}

\section{Discussions}

\paragraph{Is \model vulnerable to the selection of hyperparameters?}

There are three hyper-parameters in \model: the memory pool size $M$ that \model is able to select from; the selected memory size $m$ that is used in the forward process; and the input token size $n$ that is involved in both backward and forward process.

\begin{figure}[!htb]
    \begin{tikzpicture}[scale=0.9]
      \begin{axis}[
        xlabel={Memory Pool Size $M$},
        ylabel={Perplexity},
        xlabel style={font=\footnotesize},
        ylabel style={font=\footnotesize},
        yticklabel style={font=\footnotesize}, 
        xticklabel style={font=\footnotesize}, 
        ylabel near ticks,
        legend style={at={(1,0)},anchor=south east,mark size=2pt,font=\scriptsize,
        inner xsep=1pt, inner sep=0pt},
        legend cell align={left},
        xmin=175, xmax=625,
        ymin=23.85, ymax=24.25,
        xtick={200,300,400,500,600},
        ytick={23.9, 24.0, 24.1, 24.2}, % Here you can either remove the numbers or adjust to show fewer values
        grid = major,
        major grid style={dotted,gray},
        height=4.5cm,
        width=0.48\textwidth,
        enlarge x limits=0.05,
        ]

        % Adding the horizontal line at y=24.17
        \addplot[color=blue!50!white, solid, mark options={solid, fill=blue!50!white, draw=none}, mark=*, line width=1pt, mark size=2pt, visualization depends on=\thisrow{alignment} \as \alignment,
        every node near coord/.style={anchor=\alignment, font=\footnotesize} ] 
        table [meta index=2]  {
            x   y       label   alignment
            200 24.17   24.17   -50
            250 24.17   24.17   -50
            300 24.17   24.17   -50
            350 24.17   24.17   -50
            400 24.17   24.17   -50
            450 24.17   24.17   -50
            500 24.17   24.17   -50
            550 24.17   24.17   -50
            600 24.17   24.17   -50
        }; %
        \addlegendentry{Transformer-XL}

        % Your original plot
        \addplot[color=red!50!white, solid, mark options={solid, fill=red!50!white, draw=none}, mark=triangle*, line width=1pt, mark size=2pt, visualization depends on=\thisrow{alignment} \as \alignment,
        every node near coord/.style={anchor=\alignment, font=\footnotesize} ] 
        table [meta index=2]  {
            x   y       label   alignment
            200 24.17   24.17   -50
            250 24.04   24.04   -30
            300 23.99   23.99   -50
            350 23.98   23.98   -10
            400 23.98   23.98   -50
            450 24.00   24.00   -50
            500 24.01   24.01   -50
            550 24.03   24.03   -50
            600 24.05   24.05   -50
        }; %
        \addlegendentry{\model}

      \end{axis}

  \end{tikzpicture}
  \vspace{-2mm}
  \caption{Ablation study on memory pool size $M$ when we fix $m$=200 and $n$=64.}
  \vspace{-4mm}
  \label{fig:ablation_study_on_M} 

\end{figure}
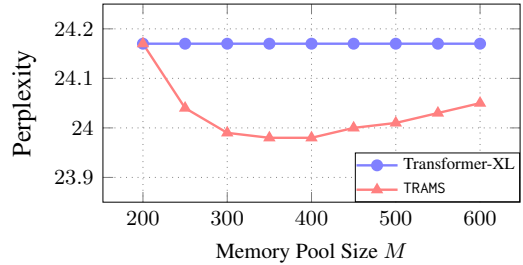

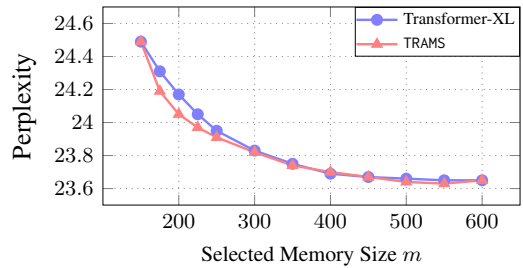
\begin{figure}[!htb]
    \begin{tikzpicture}[scale=0.9]
      \begin{axis}[
        xlabel={Selected Memory Size $m$},
        ylabel={Perplexity},
        xlabel style={font=\footnotesize},
        ylabel style={font=\footnotesize},
        yticklabel style={font=\footnotesize}, 
        xticklabel style={font=\footnotesize}, 
        ylabel near ticks,
        legend style={at={(1,1)},anchor=north east,mark size=2pt,font=\scriptsize,
        inner xsep=1pt, inner sep=0pt},
        legend cell align={left},
        xmin=125, xmax=625,
        ymin=23.50, ymax=24.70,
        xtick={200,300,400,500,600},
        ytick={23.60, 23.80, 24.00, 24.20, 24.40, 24.60},
        grid = major,
        major grid style={dotted,gray},
        height=4.5cm,
        width=0.48\textwidth,
        enlarge x limits=0.05,
        ]

        % Adding the horizontal line at y=24.17
        \addplot[color=blue!50!white, solid, mark options={solid, fill=blue!50!white, draw=none}, mark=*, line width=1pt, mark size=2pt, visualization depends on=\thisrow{alignment} \as \alignment, point meta=explicit symbolic,
        every node near coord/.style={anchor=\alignment, font=\footnotesize} ] 
        table [meta index=2]  {
            x   y       label   alignment
            150 24.49   24.49   -50
            175 24.31   24.31   -30
            200 24.17   24.17   -50
            225 24.05   24.05   -10
            250 23.95   23.95   -50
            300 23.83   23.83   -50
            350 23.75   23.75   -50
            400 23.69   23.69   -50
            450 23.67   23.67   -50
            500 23.66   23.66   -50
            550 23.65   23.65   -50
            600 23.65   23.65   -50
        }; %
        \addlegendentry{Transformer-XL}

        % Your original plot
        \addplot[color=red!50!white, solid, mark options={solid, fill=red!50!white, draw=none}, mark=triangle*, line width=1pt, mark size=2pt, visualization depends on=\thisrow{alignment} \as \alignment, point meta=explicit symbolic,
        every node near coord/.style={anchor=\alignment, font=\footnotesize} ] 
        table [meta index=2]  {
            x   y       label   alignment
            150 24.49   24.49   -50
            175 24.19   24.19   -30
            200 24.05   24.05   -50
            225 23.97   23.97   -10
            250 23.91   23.91   -50
            300 23.82   23.82   -50
            350 23.74   23.74   -50
            400 23.70   23.70   -50
            450 23.67   23.67   -50
            500 23.64   23.64   -50
            550 23.63   23.63   -50
            600 23.65   23.65   -50
        }; %
        \addlegendentry{\model}

      \end{axis}

  \end{tikzpicture}
  \vspace{-2mm}
  \caption{Ablation study on selected memory size $m$ when we fix $M$=600 and $n$=64.}
  \vspace{-4mm}
  \label{fig:ablation_study_on_m} 

\end{figure}

\begin{figure}[!htb]
    \begin{tikzpicture}[scale=0.9]
      \begin{axis}[
        xlabel={Input Length $n$},
        ylabel={Perplexity},
        xlabel style={font=\footnotesize},
        ylabel style={font=\footnotesize},
        yticklabel style={font=\footnotesize}, 
        xticklabel style={font=\footnotesize}, 
        ylabel near ticks,
        legend style={at={(1,1)},anchor=north east,mark size=2pt,font=\scriptsize,
        inner xsep=1pt, inner sep=0pt},
        legend cell align={left},
        xmin=10, xmax=130,
        ymin=23.80, ymax=24.40,
        xtick={16,32,64,128},
        ytick={23.80, 23.90, 24.00, 24.10, 24.20, 24.30, 24.40},
        grid = major,
        major grid style={dotted,gray},
        height=4.5cm,
        width=0.48\textwidth,
        enlarge x limits=0.05,
        ]

        % Adding the horizontal line at y=24.17
        \addplot[color=blue!50!white, solid, mark options={solid, fill=blue!50!white, draw=none}, mark=*, line width=1pt, mark size=2pt, visualization depends on=\thisrow{alignment} \as \alignment, point meta=explicit symbolic,
        every node near coord/.style={anchor=\alignment, font=\footnotesize} ] 
        table [meta index=2]  {
            x   y       label   alignment
            16 24.30   24.30   -50
            32 24.25   24.25   -30
            64 24.17   24.17   -50
            128 24.04   24.04   -10
        }; %
        \addlegendentry{Transformer-XL}

        % Your original plot
        \addplot[color=red!50!white, solid, mark options={solid, fill=red!50!white, draw=none}, mark=triangle*, line width=1pt, mark size=2pt, visualization depends on=\thisrow{alignment} \as \alignment, point meta=explicit symbolic,
        every node near coord/.style={anchor=\alignment, font=\footnotesize} ] 
        table [meta index=2]  {
            x   y       label   alignment
            16 24.06   24.06   -50
            32 24.03   24.03   -30
            64 23.98   23.98   -50
            128 23.90   23.90   -10
        }; %
        \addlegendentry{\model}

      \end{axis}

  \end{tikzpicture}
  \vspace{-2mm}
  \caption{Ablation study on target length $n$ when we fix $M$=400 and $m$=200.}
  \label{fig:ablation_study_on_n} 

\end{figure}
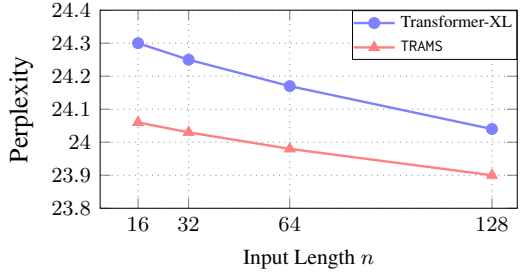

From the ablation study on $M$, Figure \ref{fig:ablation_study_on_M} suggests an optimal range between 300 to 400 for the memory pool size. Beyond this range, enlarging the memory pool often leads to the selection of irrelevant tokens, deteriorating our performance. Regarding $m$, Figure \ref{fig:ablation_study_on_m} indicates that \model witnesses a substantial drop in perplexity when the memory size selected is about 25\%. Selecting a larger portion does not yield further improvement. This is consistent with Figure \ref{fig:correlation}, where \model excels by concentrating on the top 10\% of results. Lastly, in the study on $n$, Figure \ref{fig:ablation_study_on_n} shows that as the target token length decreases, the efficacy of memory selection improves.

\paragraph{What is the inference cost compared to Transformer-XL?} 

Since there is no training part in our model, we focus on discussing the inference cost. Compared with Transformer-XL, our model requires storing a larger memory pool to do memory selection. Therefore, the memory cost of our method would be larger. When it comes to timing cost, our model has an additional memory token norm computation memory sorting operations, and memory selection operations for each layer. These extra operations require extra inference time. Table \ref{tab:infer_cost} shows the GPU memory cost and wall-clock time for the Transformer-XL baseline and our model. Our model requires slightly more GPU memory usage and around 50\% additional inference time for memory selection.

\begin{table}
\centering
\resizebox{0.48\textwidth}{!}{
\begin{tabular}{lcc}
\toprule[1.0pt]
\textbf{Model}  & \textbf{Peak GPU Mem} (MB)  & \textbf{Wall-clock Time} (s) \\
\midrule[1.0pt]
Transformer-XL & 3529 & 33.27 \\
\model & 3719 & 49.55 \\
\bottomrule[1.0pt]
\end{tabular}}
\caption{Results on GPU peak memory usage and wall-clock inference time on \wt. }
\vspace{-4mm}
\label{tab:infer_cost}
\end{table}

\paragraph{How does \model benefit from memory selection?} 

Memory selection helps the model pick tokens with higher attention scores with the queries, thus increasing the average memory utilization. Quantitatively, our method improves the average attention probability by \textbf{24.25}\% for the same size of memory compared with Transformer-XL.

\begin{figure}[t]
    \begin{tikzpicture}[scale=0.98]
      \begin{axis}[
        xlabel={Layer Number},
        ylabel={Perplexity},
        xlabel style={font=\footnotesize},
        ylabel style={font=\footnotesize},
        yticklabel style={font=\footnotesize}, 
        xticklabel style={font=\footnotesize}, 
        ylabel near ticks,
        legend style={at={(0,0)},anchor=south west,mark size=2pt,font=\scriptsize,
        inner xsep=1pt, inner sep=0pt},
        legend cell align={left},
        xmin=1, xmax=16,
        ymin=24.0, ymax=24.25,
        xtick={1, 4, 7, 10, 13, 16},
        ytick={24.0, 24.1, 24.2},
          grid = major,
          major grid style={dotted,gray},
          height=4.5cm,
          width=0.47\textwidth,
          enlarge x limits=0.05,
          ]
          
          \addplot[color=blue!50!white, solid, mark options={solid, fill=blue!50!white, draw=none}, mark=*, line width=1pt, mark size=2pt] 
          table [meta index=2]  {
            x   y       label   alignment
            1	24.17	24.17		-50
            2	24.16	24.16		-50
            3	24.17	24.17		-50
            4	24.16	24.16		-50
            5	24.16	24.16		-50
            6	24.15	24.15		-50
            7	24.17	24.17		-50
            8	24.20	24.20		-50
            9	24.16	24.16		-50
            10	24.17	24.17		-50
            11	24.18	24.18		-50
            12	24.15	24.15		-50
            13	24.07	24.07		-50
            14	24.14	24.14		-50
            15	24.13	24.13		-50
            16	24.10	24.10		-50
            }; %
            \addlegendentry{ w/ memory selection on layer $i$}

            \addplot[color=red!50!white, solid, mark options={solid, fill=red!50!white, draw=none}, mark=*, line width=1pt, mark size=2pt] 
          table [meta index=2]  {
            x   y       label   alignment
            1	24.17	24.17		-50
            2	24.17	24.17		-50
            3	24.17	24.17		-50
            4	24.17	24.17		-50
            5	24.17	24.17		-50
            6	24.17	24.17		-50
            7	24.17	24.17		-50
            8	24.17	24.17		-50
            9	24.17	24.17		-50
            10	24.17	24.17		-50
            11	24.17	24.17		-50
            12	24.17	24.17		-50
            13	24.17	24.17		-50
            14	24.17	24.17		-50
            15	24.17	24.17		-50
            16	24.17	24.17		-50
            }; %
            \addlegendentry{ w/o memory selection}
      \end{axis}

  \end{tikzpicture}
  \vspace{-2mm}
  \caption{Ablation Study on Layer-wise Importance on \wt.}
  \vspace{-2mm}
  \label{fig:layer_ablation} 

\end{figure}
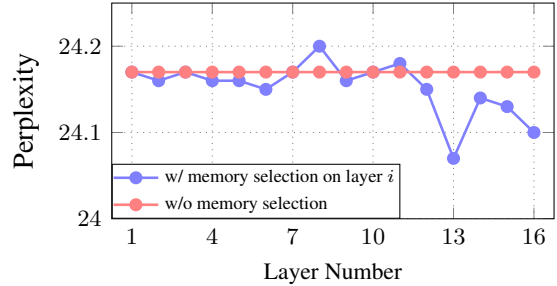

\paragraph{Does each layer hold the same importance?}
Based on Figure~\ref{fig:layer_ablation}, we show the ablation study when applying memory selection on each layer while remaining other layers the same. There is an observable drop when we apply the memory selection on the deeper layers starting from Layer 13 while we do not observe a clear influence when applying memory selection on shallow layers.

\section{Case Study}
To have an understanding of what kind of context should be selected, we provide one example case to understand specifically what kind of tokens in the memory would be selected. Based on Table \ref{tab:case_study}, we can see that most of the selected memory tokens are low-frequency words. Those low-frequency words like ``\texttt{John}" in the memory would be beneficial for the prediction of ``\texttt{John}" in the target sequence.

\begin{table}
    \centering
    \begin{tabular}{p{0.45\textwidth}}
        \toprule
        \footnotesize
        \textbf{Memory Sequence Segment} \\
        \midrule 
        \footnotesize
        \texttt{...Simon \colorbox{red!30!white}{Stephens}, which was performed in 2001 at the Royal Court Theatre. He had a guest role in the television series Judge \colorbox{red!30!white}{John} Deed in 2002. In 2004 \colorbox{red!30!white}{Boulter} landed a role as "Craig" in the episode "Teddy's Story" of the television series The Long Firm; he starred alongside actors Mark \colorbox{red!30!white}{Strong} and Derek Jacobi. He was cast in the 2005 theatre productions of the Philip Ridley play Mercury Fur, which was performed at the Drum Theatre in Plymouth and the <unk> Chocolate Factory in London. He was directed by \colorbox{red!30!white}{John} Tiffany and starred alongside Ben \colorbox{red!30!white}{Whishaw}, Shane Zaza, Harry Kent, Fraser Ayres, Sophie Stanton, and Dominic Hall. <eos> In 2006, Boulter starred alongside \colorbox{red!30!white}{Whishaw} in the play Citizenship written by Mark \colorbox{red!30!white}{Ravenhill}...} \\
        \midrule 
        \footnotesize
        \textbf{Target Sequence Segment} \\
        \midrule
        \footnotesize
        \texttt{He appeared in the television series Judge \colorbox{blue!30!white}{John} Deed in 2002 ...}\\
        \bottomrule
    \end{tabular}
    \caption{Case Study for memory selection from \wt. \colorbox{red!30!white}{\texttt{text}} indicates that this word in memory sequence is selected and used in the forward pass. \colorbox{blue!30!white}{\texttt{text}} indicates that this word in the target sequence benefits from the memory.}
    \vspace{-4mm}
    \label{tab:case_study}
\end{table}

\section{Conclusion}
In this work, we formulate the problem of memory selection in transformer architecture and reformulate the attention calculation process to obtain our self-defined queries and keys. After that, we propose a query-independent metric that utilizes memory hidden states to implement a training-free memory selector. Our experiments indicate that this method offers a simple yet effective means of identifying valuable memory tokens. Exploring optimal memory selection strategies for large language models is a promising avenue for future research. Additionally, integrating trainable parameters into these models as memory selectors presents another exciting direction for future work.

\section*{Limitations}
Our study has a couple of main limitations. First, we are currently focusing on the Transformer-XL architecture, but there are many other models with different sizes we haven't tried. It indicates that our findings could be limited to typical transformer architecture. Second, our method has many hyperparameters including $M$, $m$, and $n$. Adjusting them can greatly change how well our model works. A careful calibration is thus necessary, and one must tread cautiously to strike a balance and achieve the desired performance, which could be time-consuming and computationally expensive.

\section*{Ethics Statement}
There are no recognized potential risks.

\bibliography{anthology,custom}
\bibliographystyle{acl_natbib}

\appendix

\section{Dataset and Evaluation Metrics}
\label{sec:appendix3}
%We conduct experiments on three different datasets including word-level Wikitext-103 \citep{wikitext103} and character-level enwik8 and text8 dataset \citep{enwik8}.
\noindent \textbf{WikiText-103}~\citep{wikitext103} is a commonly used word-level language modeling benchmark. It has an average length of 3.6 thousand tokens per article and includes 28 thousand Wikipedia articles. This word-level dataset has a vocabulary size of around 260K. We use the same data pre-processing setting in \citet{transformer-xl} for this dataset. We use $\texttt{perplexity}$ as our metric.

\noindent \textbf{Enwik8}~\citep{enwik8} is a character-level language modeling benchmark. This dataset contains 100M unprocessed Wikipedia characters. The train set, dev set, and test set include 80M, 10M, and 10M characters separately. \enwik has no pre-processing stage and is directly used. $\texttt{bpc}$ (bit per character) is defined as an evaluation metric and we report results on both the dev set and test set.

\section{Training Configurations}
\label{sec:appendix}
Since we do inference experiments based on a trained model, we separately train two Transformer-XL models for \wt and \enwik. For the training stage, we use Adam~\citep{adam} to optimize with a batch size=60, learning rate=2.5e-4, target length=150, memory length=150, and a cosine scheduler without warmup steps. \par
When it comes to a different dataset, we use different Transformer-XL architecture. For \wt, we use a 16-layer transformer architecture with 10 heads, 410 hid dim, 0.1 dropout ratio, 0.0 attention dropout ratio, 2100 inner dim, and adaptive softmax mechanism. For \enwik, we propose a 12-layer transformer architecture with 8 heads, 512 hid dim, 0.1 dropout ratio, 0.0 attention dropout ratio, and 2048 inner dim. Both models are trained for 350K steps.\par 
A batch size=10 and target length=150 are fixed for all inference experiments to avoid unfair comparison. All experiments including training and inference are conducted using 4 2080Ti GPUs. It takes 280 GPU hours to train the \enwik model checkpoint. It takes 61 GPU hours to train the \wt model checkpoint. \par

\section{Relative Position Embedding}
\label{sec:appendix2}
Concerning positional encodings, we maintain the same results with Transformer-XL. The positional encodings include learnable parameters of $\boldsymbol{R}_{i-j}$, $\boldsymbol{u}$, and $\boldsymbol{v}$. Typically, $\boldsymbol{R}_{i-j}$ is derived from a learnable $r$ network included in the model. The advantage of using this design when computing the attention score is that it avoids temporal confusion caused by indexing the same position and considers the relative distance between two tokens. The formula for attention score calculation with relative position embedding can be written as:
\begin{align}\label{equ:xl-rpe}
    \boldsymbol{A}^{{xl}}_{i,j} &= \boldsymbol{X}_i^\top \boldsymbol{W}_q^\top \boldsymbol{W}_{k}^E \boldsymbol{X}_j + \boldsymbol{X}_i^\top \boldsymbol{W}_q^\top \boldsymbol{W}_{k}^R \boldsymbol{R}_{i-j} \nonumber \\
    &+ \boldsymbol{u}^\top \boldsymbol{W}_{k}^E \boldsymbol{X}_j + \boldsymbol{v}^\top \boldsymbol{W}_{k}^R \boldsymbol{R}_{i-j}
\end{align}
Moreover, after doing ablation studies on relative position embedding, we found that $\boldsymbol{R}_{i-j}$ contributes the most to the result and $\boldsymbol{u}$, $\boldsymbol{v}$ only has a small influence on the final performance. The existence of $\boldsymbol{R}_{i-j}$ leads to the exponentially decayed attention probability distribution related to a memory position. As a result, we base our memory selection on the $\boldsymbol{A}^{{xl}}_{i,j}$ which includes positional information instead of the pure $\boldsymbol{X}_i^\top \boldsymbol{W}_q^\top \boldsymbol{W}_{k}^E \boldsymbol{X}_j$. To be noticed, all concepts related to $\mathbf{q}\mathbf{K}$ are all equipped with position embedding instead of a simple dot product.

\end{document}